%% file: main.tex
\documentclass[lettersize,journal]{IEEEtran}
\usepackage{amsmath,amsfonts}
\usepackage{algorithm}
\usepackage{array}
\usepackage[caption=false,font=normalsize,labelfont=sf,textfont=sf]{subfig}
\usepackage{textcomp}
\usepackage{stfloats}
\usepackage{url}
\usepackage{verbatim}
\usepackage{graphicx}
\usepackage{cite}
\usepackage{siunitx}
\usepackage{xcolor}
\usepackage{longtable}
\usepackage{dsfont}
\usepackage{subfig} %[position=top]
\usepackage{multirow}
\usepackage{textcomp}
\usepackage{tabularx}
\usepackage{enumerate}
\usepackage{nameref}
\usepackage{graphicx}
\usepackage{wrapfig}
\usepackage{makecell}
\usepackage{algpseudocode}
\usepackage{mathtools}

\usepackage{comment}

\usepackage{booktabs}
\makeatletter
\def\RS#1{\zap@space#1 \@empty}
\usepackage{bbm}
\usepackage[T1]{fontenc}

\begin{document}

\title{LaplaceConfidence: a Graph-based Approach for Learning with Noisy Labels}

\author{Mingcai Chen $^1$, Yuntao Du $^1$, Wei Tang $^2$, Baoming Zhang $^1$, Hao Cheng $^1$, Shuwei Qian $^1$, Chongjun Wang $^{1*}$\thanks{$^*$Corresponding author}\\
$^1$ State Key Laboratory for Novel Software Technology at Nanjing University, Nanjing University\\
$^2$ Department of Neurology, University Medical Center Groningen, University of Groningen

        % <-this % stops a space
% <-this % stops a space
}

% The paper headers
% \markboth{IEEE Transactions on Knowledge and Data Engineering}%
% {Shell \MakeLowercase{\textit{et al.}}: A Sample Article Using IEEEtran.cls for IEEE Journals}

\IEEEpubid{0000--0000/00\$00.00~\copyright~2021 IEEE}
% Remember, if you use this you must call \IEEEpubidadjcol in the second
% column for its text to clear the IEEEpubid mark.

\maketitle

\begin{abstract}
In real-world applications, perfect labels are rarely available, making it challenging to develop robust machine learning algorithms that can handle noisy labels. Recent methods have focused on filtering noise based on the discrepancy between model predictions and given noisy labels, assuming that samples with small classification losses are clean. This work takes a different approach by leveraging the consistency between the learned model and the entire noisy dataset using the rich representational and topological information in the data. We introduce LaplaceConfidence, a method that to obtain label confidence (i.e., clean probabilities) utilizing the Laplacian energy. Specifically, it first constructs graphs based on the feature representations of all noisy samples and minimizes the Laplacian energy to produce a low-energy graph. Clean labels should fit well into the low-energy graph while noisy ones should not, allowing our method to determine data's clean probabilities. Furthermore, LaplaceConfidence is embedded into a holistic method for robust training, where co-training technique generates unbiased label confidence and label refurbishment technique better utilizes it. We also explore the dimensionality reduction technique to accommodate our method on large-scale noisy datasets. Our experiments demonstrate that LaplaceConfidence outperforms state-of-the-art methods on benchmark datasets under both synthetic and real-world noise.
% Code available at \url{https://anonymous.4open.science/r/LaplaceConfidence-07D0}.
\end{abstract}

\input{content.tex}

\section{Acknowledgements}
This paper is supported by the National Natural Science Foundation of China (Grant No. 62192783, U1811462), the Collaborative Innovation Center of Novel Software Technology and Industrialization at Nanjing University.

\bibliographystyle{IEEEtran}
\bibliography{egbib}
% \begin{thebibliography}{1}
% \bibliographystyle{IEEEtran}

% \bibitem{ref1}
% {\it{Mathematics Into Type}}. American Mathematical Society. [Online]. Available: https://www.ams.org/arc/styleguide/mit-2.pdf

% \bibitem{ref2}
% T. W. Chaundy, P. R. Barrett and C. Batey, {\it{The Printing of Mathematics}}. London, U.K., Oxford Univ. Press, 1954.

% \bibitem{ref3}
% F. Mittelbach and M. Goossens, {\it{The \LaTeX Companion}}, 2nd ed. Boston, MA, USA: Pearson, 2004.

% \bibitem{ref4}
% G. Gr\"atzer, {\it{More Math Into LaTeX}}, New York, NY, USA: Springer, 2007.

% \bibitem{ref5}M. Letourneau and J. W. Sharp, {\it{AMS-StyleGuide-online.pdf,}} American Mathematical Society, Providence, RI, USA, [Online]. Available: http://www.ams.org/arc/styleguide/index.html

% \bibitem{ref6}
% H. Sira-Ramirez, ``On the sliding mode control of nonlinear systems,'' \textit{Syst. Control Lett.}, vol. 19, pp. 303--312, 1992.

% \bibitem{ref7}
% A. Levant, ``Exact differentiation of signals with unbounded higher derivatives,''  in \textit{Proc. 45th IEEE Conf. Decis.
% Control}, San Diego, CA, USA, 2006, pp. 5585--5590. DOI: 10.1109/CDC.2006.377165.

% \bibitem{ref8}
% M. Fliess, C. Join, and H. Sira-Ramirez, ``Non-linear estimation is easy,'' \textit{Int. J. Model., Ident. Control}, vol. 4, no. 1, pp. 12--27, 2008.

% \bibitem{ref9}
% R. Ortega, A. Astolfi, G. Bastin, and H. Rodriguez, ``Stabilization of food-chain systems using a port-controlled Hamiltonian description,'' in \textit{Proc. Amer. Control Conf.}, Chicago, IL, USA,
% 2000, pp. 2245--2249.

% \end{thebibliography}

\newpage

% \section{Biography Section}
% If you have an EPS/PDF photo (graphicx package needed), extra braces are
%  needed around the contents of the optional argument to biography to prevent
%  the LaTeX parser from getting confused when it sees the complicated
%  $\backslash${\tt{includegraphics}} command within an optional argument. (You can create
%  your own custom macro containing the $\backslash${\tt{includegraphics}} command to make things
%  simpler here.)
 
% \vspace{11pt}

% \bf{If you include a photo:}\vspace{-33pt}
% \begin{IEEEbiography}[{\includegraphics[width=1in,height=1.25in,clip,keepaspectratio]{fig1}}]{Michael Shell}
% Use $\backslash${\tt{begin\{IEEEbiography\}}} and then for the 1st argument use $\backslash${\tt{includegraphics}} to declare and link the author photo.
% Use the author name as the 3rd argument followed by the biography text.
% \end{IEEEbiography}

% \vspace{11pt}

% \bf{If you will not include a photo:}\vspace{-33pt}
% \begin{IEEEbiographynophoto}{John Doe}
% Use $\backslash${\tt{begin\{IEEEbiographynophoto\}}} and the author name as the argument followed by the biography text.
% \end{IEEEbiographynophoto}

% \clearpage

\end{document}

%% file: content.tex
\section{Introduction}
The success of deep learning relies on high-quality labeled datasets.
However, various obstacles, such as expensive labor costs, large quantities of data, or requirements for domain knowledge, could exist in the labeling process \cite{DBLP:journals/corr/abs-2011-04406,zhou2018brief}.
Consequently, fully intact labels are usually not readily available in real-world applications. 
What's more, recent studies have found that label noise can seriously damage the performance of deep models \cite{zhang2016understanding,arpit2017closer}.
Therefore, it is necessary to improve model robustness against noisy labels.

\begin{figure}
    \centering
    \includegraphics[width=\columnwidth]{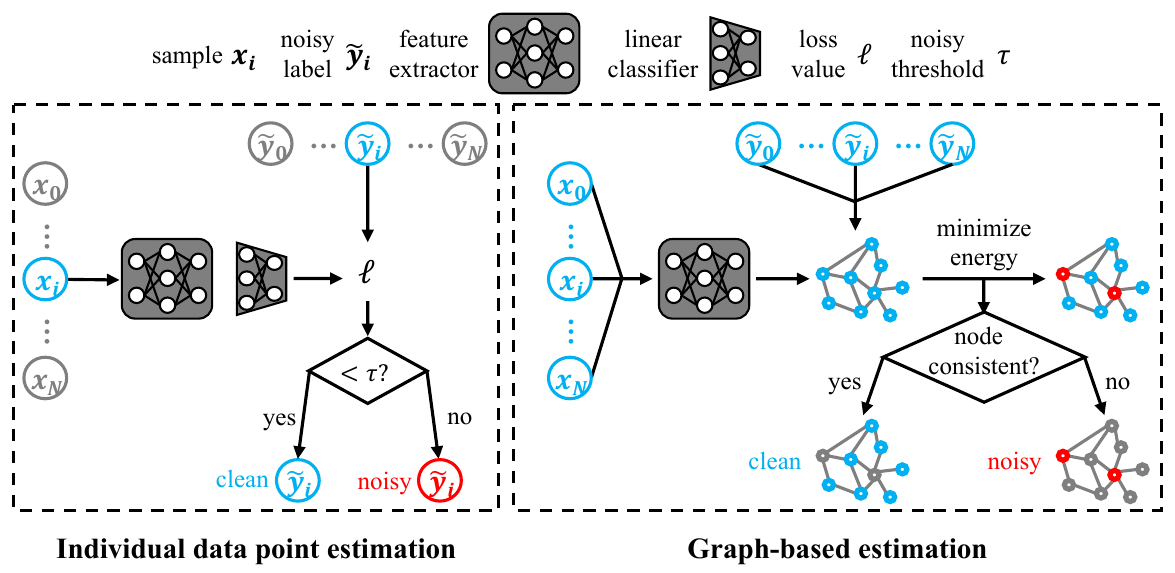} 
    \caption{%DivideMix's (a) test accuracy and (b) quality of confidence estimation during training.
    % (a) and (b) show the overlap of loss distribution on CIFAR-10 under different noise levels. 
            Conventional individual data point estimation vs. our graph-based estimation. Our graph-based method takes all samples and their topological relationship into consideration.
        }
    \label{fig_motivation}
\end{figure}

The field of Learning with Noisy Labels (LNL) encompasses a broad spectrum of algorithms. %has attracted lots of attention recently and 
Recently, a series of methods \cite{jiang2018mentornet,han2018co,li2020dividemix} has significantly improved robustness by leveraging the memorization effect, i.e., the behavior that deep models fit generalizable patterns before memorizing the noisy patterns \cite{arpit2017closer}.
Networks tend to present smaller losses on clean samples and vice versa.
Therefore, the clean probability of given labels, i.e., label confidence, can be estimated according to per-sample loss, as shown in the left of Fig. \ref{fig_motivation}. %, which is commonly referred to as ``small-loss trick.''
For example, DivideMix \cite{li2020dividemix} uses Gaussian Mixture Model (GMM) to dynamically find the loss threshold $\tau$ for sample selection.
It then trains the model using the supervised training signals from clean labels and self-supervised training signals from noisy examples.
However, there is a growing awareness of the defects in the loss-based criteria \cite{song2020learning,wu2020topological,wu2021ngc}: 
The loss distribution of true-labeled and false-labeled samples always overlap,
especially when the noise rate is heavy or there exist some hard-to-learn samples \cite{song2020learning,DBLP:journals/corr/abs-2108-11096}.

\IEEEpubidadjcol
The central tenets of our work are twofold:
1). the previous observation \cite{DBLP:conf/cvpr/OrtegoAAOM21, DBLP:conf/cvpr/IscenVAS22} that suggests overfitting to noisy labels occurs less in hidden representations, 
2). the label confidence can be better estimated based on relationship of the learned representations than individuals of those. 
To support our claims, we first verify the previous observation that, even when trained on noisy datasets, samples tend to have same-class neighbors in the learned feature space (see Section \ref{sec:cifar} for our analysis).
On this basis, we can relate how possible a sample is mislabeled to how many different-class neighbors it has  \cite{song2019selfie,DBLP:conf/cvpr/OrtegoAAOM21,DBLP:conf/icml/BahriJG20}.
However, the mislabeled data points could also affect and bias the prediction of their neighbors.
To address this issue, we aim to update labels to reach a global optimal estimation where label consistency of all data points has converged to a stable state. 
To mathematically model this problem, we borrow the concept of Laplacian energy in graph theory \cite{newman2004detecting,DBLP:conf/icml/ZhuGL03,DBLP:journals/corr/abs-2106-04527}, which qualifies the label consistency between each node and its neighbors in the connection structure.
We estimate label confidence by solving the graph Laplacian minimization problem.
Specifically, a graph is first built using the features extracted from the model, where nodes represent samples and the edges are determined by the similarity between pairs of samples.
On the graph, the noisy labels of all samples are optimized so that the disagreement between neighbors is minimized.
After this process, the clean nodes tend to remain unaffected, while the noisy labels would be changed because they are inconsistent with a low-energy graph whose neighbors have similar labels.

In conclusion, we propose a novel label confidence estimation method named LaplaceConfidence, fully utilizing the learned representations and their topological relationship.
We embed it into a holistic method by combining it with other techniques, including co-training, label refurbishment \cite{reed2014training,arazo2019unsupervised}, and data augmentation.
Given that the real-world applications of LNL often involve large-size models or datasets \cite{li2017webvision}, we also investigate the role of the dimensionality reduction technique in the scalability of our method.
The main contributions of this work are as follows:
\begin{itemize}
    \item We propose a feature-based confidence estimation method that obtains optimal confidence estimation based on the topological information of samples. Besides, LaplaceConfidence is integrated with other techniques to form a holistic approach for learning with noisy labels.
    \item To make our method suitable for large-scale networks or large noisy datasets in real-world applications, we investigate the role of dimensionality reduction before using the feature representations to construct the graph. We find that it can significantly accelerate our method without compromising performance. In some cases, it can even improve robustness.
    \item We demonstrate that LaplaceConfidence outperforms previous classification-loss-based estimation methods.
    Our method achieves state-of-the-art results on standard LNL benchmarks, namely CIFAR-10, CIFAR-100 with synthetic label noise and real-world noisy dataset Mini-WebVision.
    We also systematically study the components of LaplaceConfidence to examine their effectiveness.
\end{itemize}

\section{Related work}
This section first introduces the taxonomy of label noise.
Then we categorize some recent LNL algorithms into two groups, namely classification-loss-based and feature-based, to set the stage for LaplaceConfidence.
We also introduce related semi-supervised learning works.

\subsection{Taxonomy of Label Noise}
Formally, let $\tilde{y}$, $y$ denote the noisy label and underlying true label, respectively. 
The distribution of noisy labels is affected by the dependency between data features and class labels: $p(\tilde{y}=j\mid x,y)$.
Based on this, some \cite{liu2015classification,sukhbaatar2014training,chen2015webly,patrini2017making} assume an instance-independent label noise model:
\begin{equation}
\begin{aligned}
\text{p}(\tilde{y}=j\mid x,y)&=\sum_{i=1}^C \text{p}(\tilde{y}=j,y=i \mid x)\\
&=\sum_{i=1}^{C}\text{p}(\tilde{y}=j\mid y=i)\text{p}(y=i\mid x)
\end{aligned}
\end{equation}
where $C$ is the number of classes.
$\text{p}(\tilde{y}=j\mid y=i)$ is the noise model, i.e., the probability of the sample in class $i$ being corrupted to class $j$.
The second equation holds because the label noise is assumed to be independent of input features.
% We can use a transition matrix $T\in [0,1]^{C\times C}$ to model such noise, where $\mathbbm{1}_CT=\mathbf{1}$.

However, the instance-independent noise model is unrealistic. 
For example, in some real-world datasets, an unrecognizable image is more likely to be mislabeled. 
Therefore, the corruption probability should be dependent on the data features, i.e., instance-dependent label noise \cite{menon2018learning,cheng2020learning}.
Explicitly modeling such noise is challenging.
Some recent methods \cite{li2020dividemix,nishi2021augmentation,wu2021ngc} leverage confidence-based sample selection and semi-supervised learning, achieving significant improvement.

\subsection{Classification-loss-based LNL}
In Mentornet \cite{jiang2018mentornet}, a pre-trained network selects samples with small losses to train a student network.
Co-teaching \cite{han2018co}, on the other hand, maintains two equivalent models, and each selects small-loss samples for its peer.
Co-teaching+ \cite{yu2019does} extends Co-teaching by further filtering agreed predictions, so the two base models would not collapse into a consensus.
ITLM \cite{shen2019learning} aims to optimize a trimmed loss by jointly choosing a fraction of samples and updating the model on it.
INCV \cite{chen2019understanding} selects clean samples from the noisy ones at every round via a cross-validation process.
Then it deploys the co-teaching training schema.
SELFIE \cite{song2019selfie} refurbishes small-loss samples with the most frequently predicted labels in previous training epochs.
\cite{arazo2019unsupervised} fits the distribution of pre-sample loss on a Beta Mixture Model (BMM).
Then the produced probability is used as the coefficient of the bootstrapping loss \cite{reed2014training}.
\cite{lyu2019curriculum} designs a surrogate loss of the robust 0-1 loss and uses it for clean sample selection.
DivideMix \cite{li2020dividemix} uses a GMM to model the loss distribution and identify the noise.
Possible incorrect labels are eliminated before performing a semi-supervised learning algorithm named MixMatch \cite{berthelot2019mixmatch} to fully utilize all data.
Different from these, our method investigates the LNL problem from a latent representational and topological perspective.

\subsection{Feature-based LNL}
There are several recent LNL methods \cite{song2019selfie,DBLP:conf/cvpr/OrtegoAAOM21,DBLP:conf/icml/BahriJG20} based on latent feature representation. 
Dimensionality-Driven Learning (D2L) \cite{ma2018dimensionality} adopts a label refurbishment framework and backpropagates the loss for a linear combination of predictions and noisy labels.
The method chooses an optimal weight, i.e., label confidence, for the combination such that the increase of local intrinsic dimensionality \cite{houle2017local} is prevented.
TopoFilter \cite{wu2020topological} filters noise according to the topological relationship between samples.
It constructs a $k$-NN graph and treats nodes in the largest connected component for each class as clean samples.
Multi-Objective Interpolation Training \cite{DBLP:conf/cvpr/OrtegoAAOM21} identifies noise by comparing the predictions of samples and those of their neighbors before correcting the wrong labels for training.
Noisy Graph Cleaning (NGC) \cite{wu2021ngc} considers a new problem setup: learning with open-world noisy data.
The method constructs a graph and performs label propagation to obtain pseudo-labels.
Then it selects clean samples using the largest connected component within each class.
Besides, it adopts contrastive learning at a sub-graph level.
Neighbor Consistency Regularization (NCR) \cite{DBLP:conf/cvpr/IscenVAS22} deploys a simple regularization loss term for robust training, encouraging examples with similar feature representations to have similar predictions.
However, we suggest that the topological information can be further utilized by optimizing the whole dataset's topological structure.
The major distinction of LaplaceConfidence is that it reaches an optimal estimation by minimizing the graph Laplacian energy.

\subsection{Semi-Supervised Learning}
Semi-supervised learning aims to leverage both labeled data and unlabeled data.
Recent LNL methods attempt to convert the LNL problem into a semi-supervised learning problem by removing some possible noisy labels and utilizing powerful semi-supervised learning techniques.
For instance, DivideMix's success can be largely attributed to the deployment of MixMatch.
Our idea of constructing a graph using data points' representations is inspired by the semi-supervised learning method LaplaceNet \cite{DBLP:journals/corr/abs-2106-04527}, which assigns pseudo-labels to unlabeled data using the label propagation algorithm.
One difference between LaplaceNet and our method is that the former employs propagated labels for training, while we utilize the resulting graph for label confidence estimation.
One may wonder why LaplaceConfidence does not directly use the refined labels as training targets.
We find that, unlike in semi-supervised learning, this would yield suboptimal results in LNL.
We note that obtaining pseudo-labels in each iteration (instead of per epoch) using the latest model leads to less noisy training targets, which is essential to an LNL method. 
Therefore, LaplaceConfidence estimates label confidence in every epoch and generates new training targets in every iteration.

\section{Method}
\label{sec:method}
\subsection{Problem Formulation}
Different from the standard supervised learning, only a noisy training dataset $\mathcal{\tilde{D}}=\{\boldsymbol x_i,\tilde{\boldsymbol{y}}_i\}_{i=1}^{N}$ is available in LNL, where $\boldsymbol x$ is the input feature and $\tilde{\boldsymbol{y}}\in \{0,1\}^C$ are the one-hot noisy label vector in $C$-class. $N$ is the number of training samples.
LNL is to train a robust model, which can be viewed as a composition of a feature extractor $g(\cdot)$ and a linear classifier $f(\cdot)$.
The performance is evaluated on a clean test dataset.

\subsection{LaplaceConfidence}
\label{method_lc}

To model the topological relationship of samples, we utilize the graph structure.
Specifically, an undirected $k$-NN graph is first constructed using penultimate layer features of all training samples $\{\boldsymbol v_1=g(\boldsymbol x_1),\dots,\boldsymbol v_N=g(\boldsymbol x_N)\}$. The weighted adjacency matrix $A$: %with the penultimate layer features from a pruned network
\begin{equation}
\label{E weighted adjacency matrix}
    A_{ij}=\left\{ \begin{array}{ll}
         \langle \boldsymbol v_i,\boldsymbol v_j\rangle , & \text{if}\, i\in \mathcal{N}_k(j)\\%\,i\neq j\,\text{and}\,
         0, & \text{otherwise}
    \end{array}
    \right.
\end{equation}
where $\langle \cdot,\cdot \rangle $ is the inner product. 
$\mathcal{N}_k$ denotes the $k$ nearest neighbors.
To balance the influence of samples with different numbers of neighbors, the diagonal degree matrix $D$ is used to normalize $A$: 
\begin{equation}
\begin{aligned}
\label{E normalized adjacency matrix}
    \bar{A} = &D^{-1/2}AD^{-1/2} \\
    D = & \text{diag}(A\mathbbm{1}_N)
\end{aligned}
\end{equation}
where diag$(\cdot)$ is a diagonal matrix whose diagonal consists of the vector in the bracket.

A graph formed from a clean dataset should have low graph Laplacian energy because samples are likely to agree with their neighbors' labels.
Our obtained graph is not the case due to the inconsistency between the learned features and the noisy labels. 
Therefore, we can identify noisy nodes that cause such inconsistency according to which nodes should be changed for a low-energy graph structure.
The graph Laplacian energy over the label distribution is minimized to obtain the clean graph structure.
\begin{equation}
\label{E graph laplacian}
    \mathcal{Q}(\bar{Y})=\frac{1}{2}\sum^N_{i,j=1}\bar{A}_{ij} \left\Vert \frac{\bar{\boldsymbol{y}}_i}{\sqrt{D_{ii}}} - \frac{\bar{\boldsymbol{y}}_j}{\sqrt{D_{jj}}} \right\Vert ^2 +\frac{\mu}{2}\sum_{i=1}^N \Vert \bar{\boldsymbol{y}}_i-\tilde{\boldsymbol{y}}_i\Vert^2
\end{equation}
where the refined label distribution $\bar{Y}=[\bar{\boldsymbol{y}}_1,\cdots,\bar{\boldsymbol{y}}_N]\in \mathbb{R}^{N\times C}$. 
The second term is a fidelity term that avoids the refined labels changing too much from the original labels.
$\mu$ is the coefficient that balances between the node's neighborhoods and itself. 
Minimizing $\mathcal{Q}$ is a typical convex optimization problem.
To side-step the calculation of matrix inverse, we use the conjugate gradient method to solve the linear system $(I-(1+\mu)^{-1}\bar{A})\bar{Y}=\tilde{Y}$.
The calculation follows the prevalent practice, so we don't elaborate on the details here. 
Finally, a global optimal label distribution is obtained.

Having the refined label distribution $\bar{Y}$, there are many ways to map it to clean probability, e.g., calculating the cross-entropy between the original labels $\mathrm{H}(\bar{\boldsymbol{y}},\tilde{\boldsymbol{y}})$ and then put it in previous GMM framework.
We find that simply using the probability on the original class as label confidence $w=\bar{\boldsymbol{y}}[\tilde{y}]$ yields good results. 

\subsection{The Overall Training Process}

\begin{algorithm}
    \caption{Training schema for LaplaceConfidence} 
    \label{alg}
    \begin{algorithmic}[1] 
    \item[] \textbf{Input}: Noisy dataset $\tilde{\mathcal{D}}=\{(\boldsymbol x_i,\tilde{\boldsymbol y}_i)\}^{N}_{i=1}$,  \# total rounds for training $R$,  \# rounds for warm-up $R_{\text{warm}}$, \# iterations of every training round $I$. 
    \item[] \textbf{Output}: Trained model.
    \State Randomly initialize two models
    \For{$r=1$ to $R$} 
        \For{$m=0$ to $1$}
            % \Comment{train two models in turn}
            \If{$r<R_{\text{warm}}$} %\Comment{supervised training to warm the models}
                \State Train model $m$ over $\tilde{\mathcal{D}}$ for one epoch 
            \Else
                \State Extract features and construct graph following Eq.(\ref{E weighted adjacency matrix}) and Eq.(\ref{E normalized adjacency matrix})
                \State Minimize Eq.(\ref{E graph laplacian}) and obtain label confidence $w$ 
                \State Refurbish labels following Eq.(\ref{E label refurbishment}) and Eq.(\ref{E coteaching})
                \State Train model $m$ over $\{\mathcal{A}(\boldsymbol x_i),\boldsymbol y^*_i\}_{i=1}^N$ for $I$ iterations
            \EndIf
        \EndFor
    \EndFor
    \end{algorithmic} 
\end{algorithm}

Though we mainly focus on label confidence estimation in this work, some other techniques are also unified to form a holistic pipeline for LNL.
The whole training schema is described in Algorithm \ref{alg}. 
% Some frequently used notations and their meanings are presented in Table \ref{table_notation}.

Specifically, the label refurbishment framework trains the model with the refurbished label $\boldsymbol y^*$, which is obtained from a convex combination of the noisy label $\tilde{\boldsymbol{y}}$ and the pseudo-label $\hat{\boldsymbol{y}}$ from the model's prediction.
\begin{equation}
\label{E label refurbishment}
    \boldsymbol y^*=w\tilde{\boldsymbol{y}}+(1-w)\hat{\boldsymbol{y}}
\end{equation}
where $w$ is the label confidence from Sec. \ref{method_lc}. 
The bigger the label confidence, the more the model fits the given label.
The smaller the label confidence, the more the optimization objective leans toward self-training.

% To improve generalization, the image augmentation method RandAugment \cite{cubuk2020randaugment} is used for training.
% It selects a set of basic image transformations to operate on input images sequentially.
% Note that LaplaceConfidence only performs basic random crop and random horizontal flip when estimating confidence and generating pseudo-labels.

Using one model's own predictions to guide its subsequent training leads to the error accumulation problem \cite{tarvainen2017mean,arazo2020pseudo}.
Co-training alleviates the problem by training two models simultaneously.
We adopt the co-training schema in DivideMix \cite{li2020dividemix}.
Specifically, two models with the same structure but different parameter initialization are maintained.
The confidence one model uses comes from its peer.
For pseudo-labeling, two models' predictions are ensembled.
Let $\text{p}_\text{model1}(\boldsymbol y\mid \boldsymbol x)$ and $\text{p}_\text{model2}(\boldsymbol y\mid \boldsymbol x)$ be the two networks' predictions, respectively. 
The pseudo-labels are generated by:
\begin{equation}
\label{E coteaching}
\begin{aligned}
&\hat{\boldsymbol{y}}=\text{Sharpen}\left(\frac{\text{p}_\text{model1}\left(\boldsymbol y\mid \alpha\left(\boldsymbol x\right)\right)+\text{p}_\text{model2}\left(\boldsymbol y\mid \alpha\left(\boldsymbol x\right)\right)}{2}\right)\\
\end{aligned}
\end{equation}
where $\alpha(\cdot)$ is a basic image augmentation function, which randomly flips and crops the input images. 
$\text{Sharpen}(\text{p}_i) = \text{p}_i^{\frac{1}{T}} \bigg/ \sum_{j=1}^{C} \text{p}_j^{\frac{1}{T}}$ reduces the entropy of the label distribution $\textbf{\text{p}} = (\text{p}_1,\dots,\text{p}_C)$ with a temperature $T$.
% The $\text{Sharpen}(\cdot)$ here is the same as in Sec. \ref{mixup} but uses a different temperature $T_{pl}$ for scaling.

We use mini-batch stochastic gradient descent algorithm for optimization.
The loss of a sample $\boldsymbol x$ in a mini-batch is the cross-entropy $\text{H}$ between the soft pseudo-labels and the predictions of the model:
\begin{equation}
    \mathcal{L}=\text{H}\left(\boldsymbol y^*,f\left(g\left(\mathcal{A}\left(\boldsymbol x\right)\right)\right)\right)
\end{equation}
where $\mathcal{A}$ is an augmentation method RandAugment \cite{cubuk2020randaugment}.
% used in many recent weakly supervised learning algorithms \cite{sohn2020fixmatch,nishi2021augmentation}.
It first randomly selects a given number of operations from a set of image transformations, including geometric and photometric transformations. 
Sequentially, these operations are applied with random magnitudes.
Its details are in the supplementary material. 
$f$ and $g$ is the linear classifier and feature extractor, respectively.

% \textbf{Implementation} We provide an effecient implementation

\section{Results and Discussion}

\subsection{Experimental Details}

\begin{table*}
    \caption{Comparison with state-of-the-art methods on CIFAR-10 and CIFAR-100 with synthetic noise. Sym. and Asym. are symmetric and asymmetric for short, respectively. The best results are indicated in bold. We include the results of AugDesc* with two different augmentation policies.}
    \label{table_CIFAR}
    \centering
    \begin{tabular}{lr|c|c|c|c|c||c|c|c|c}
        \toprule
		Dataset               &      &\multicolumn{5}{c||}{CIFAR-10}& \multicolumn{4}{c}{CIFAR-100}\\\midrule%\cmidrule(r){1-2}\cmidrule(Robust LR){3-7}\cmidrule(l){8-11}
		Noise type &      &\multicolumn{4}{c|}{Sym.}& \multicolumn{1}{c||}{Asym.} & \multicolumn{4}{c}{Sym.}\\\midrule
		\multicolumn{2}{l|}{Method/Noise ratio}        & 20\% & 50\% & 80\% & 90\% & 40\% & 20\% & 50\% & 80\% &  90\% \\ \midrule%\cmidrule(r){1-2}\cmidrule(Robust LR){3-7}\cmidrule(l){8-11}
% 		\multirow{2}{*}{Cross-Entropy}         & Best  & 86.8 & 79.4 & 62.9 & 42.7 & 85.0 & 62.0 & 46.7 & 19.9 & 10.1 \\
% 		                                       & Last  & 82.7 & 57.9 & 26.1 & 16.8 & 72.3 & 61.8 & 37.3 &  8.8 &  3.5 \\ \midrule
		\multirow{2}{*}{Bootstrap}             & Best  & 86.8 & 79.8 & 63.3 & 42.9 &  -   & 62.1 & 46.6 & 19.9 & 10.2 \\
		                                       & Last  & 82.9 & 58.4 & 26.8 & 17.0 &  -   & 62.0 & 37.9 &  8.9 &  3.8 \\ \midrule
		\multirow{2}{*}{F-correction}          & Best  & 86.8 & 79.8 & 63.3 & 42.9 & 87.2 & 61.5 & 46.6 & 19.9 & 10.2 \\
		                                       & Last  & 83.1 & 59.4 & 26.2 & 18.8 & 83.1 & 61.4 & 37.3 &  9.0 &  3.4 \\ \midrule
		\multirow{2}{*}{Co-teaching+}          & Best  & 89.5 & 85.7 & 67.4 & 47.9 &  -   & 65.6 & 51.8 & 27.9 & 13.7 \\
		                                       & Last  & 88.2 & 84.1 & 45.5 & 30.1 &  -   & 64.1 & 45.3 & 15.5 &  8.8 \\ \midrule
		\multirow{2}{*}{Mixup}                 & Best  & 95.6 & 87.1 & 71.6 & 52.2 &  -   & 67.8 & 57.3 & 30.8 & 14.6 \\
		                                       & Last  & 92.3 & 77.6 & 46.7 & 43.9 &  -   & 66.0 & 46.6 & 17.6 &  8.1 \\ \midrule
		\multirow{2}{*}{P-correction}          & Best  & 92.4 & 89.1 & 77.5 & 58.9 & 88.5 & 69.4 & 57.5 & 31.1 & 15.3 \\
		                                       & Last  & 92.0 & 88.7 & 76.5 & 58.2 & 88.1 & 68.1 & 56.4 & 20.7 &  8.8 \\\midrule
		\multirow{2}{*}{Meta-Learning}         & Best  & 92.9 & 89.3 & 77.4 & 58.7 & 89.2 & 68.5 & 59.2 & 42.4 & 19.5 \\
		                                       & Last  & 92.0 & 88.8 & 76.1 & 58.3 & 88.6 & 67.7 & 58.0 & 40.1 & 14.3 \\\midrule
		\multirow{2}{*}{M-correction}          & Best  & 94.0 & 92.0 & 86.8 & 69.1 & 87.4 & 73.9 & 66.1 & 48.2 & 24.3 \\
		                                       & Last  & 93.8 & 91.9 & 86.6 & 68.7 & 86.3 & 73.4 & 65.4 & 47.6 & 20.5 \\\midrule
		\multirow{2}{*}{DivideMix}             & Best  & 96.1 & 94.6 & 93.2 & 76.0 & 93.4 & 77.3 & 74.6 & 60.2 & 31.5 \\
				                               & Last  & 95.7 & 94.4 & 92.9 & 75.4 & 92.1 & 76.9 & 74.2 & 59.6 & 31.0 \\\midrule
	    \multirow{2}{*}{AugDesc-AutoAugment$^*$}& Best  & 96.3 &  95.4&  93.8& 91.9 &  94.6& 79.5 &  \textbf{77.2}   &  66.4   & 41.2 \\
				                               & Last  & 96.2 &  95.1&  93.6& 91.8 &  94.3& 79.2 &  \textbf{77.0}   &  66.1   & 40.9 \\\midrule 
	    \multirow{2}{*}{AugDesc-RandAugment$^*$}& Best  & 96.1 &  -   &   -  & 89.6 &  -   & 78.1 &  -   &  -   & 36.8 \\
				                               & Last  & 96.0 &  -   &   -  & 89.4 &  -   & 77.8 &  -   &  -   & 36.7 \\\midrule 
		\multirow{2}{*}{LaplaceConfidence}            & Best & \textbf{96.4} & \textbf{96.0} & \textbf{95.0} & \textbf{94.7} & \textbf{95.2} & \textbf{79.6} & 76.5 & \textbf{70.4} &  \textbf{55.2} \\
				              & Last & \textbf{96.3} & \textbf{95.8} & \textbf{94.8} & \textbf{94.6} & \textbf{94.7} & \textbf{79.3} & 75.5 & \textbf{69.4} & \textbf{44.6} \\
        \bottomrule
    \end{tabular}
\end{table*}

\begin{figure}
    \centering
    \includegraphics[width=\columnwidth]{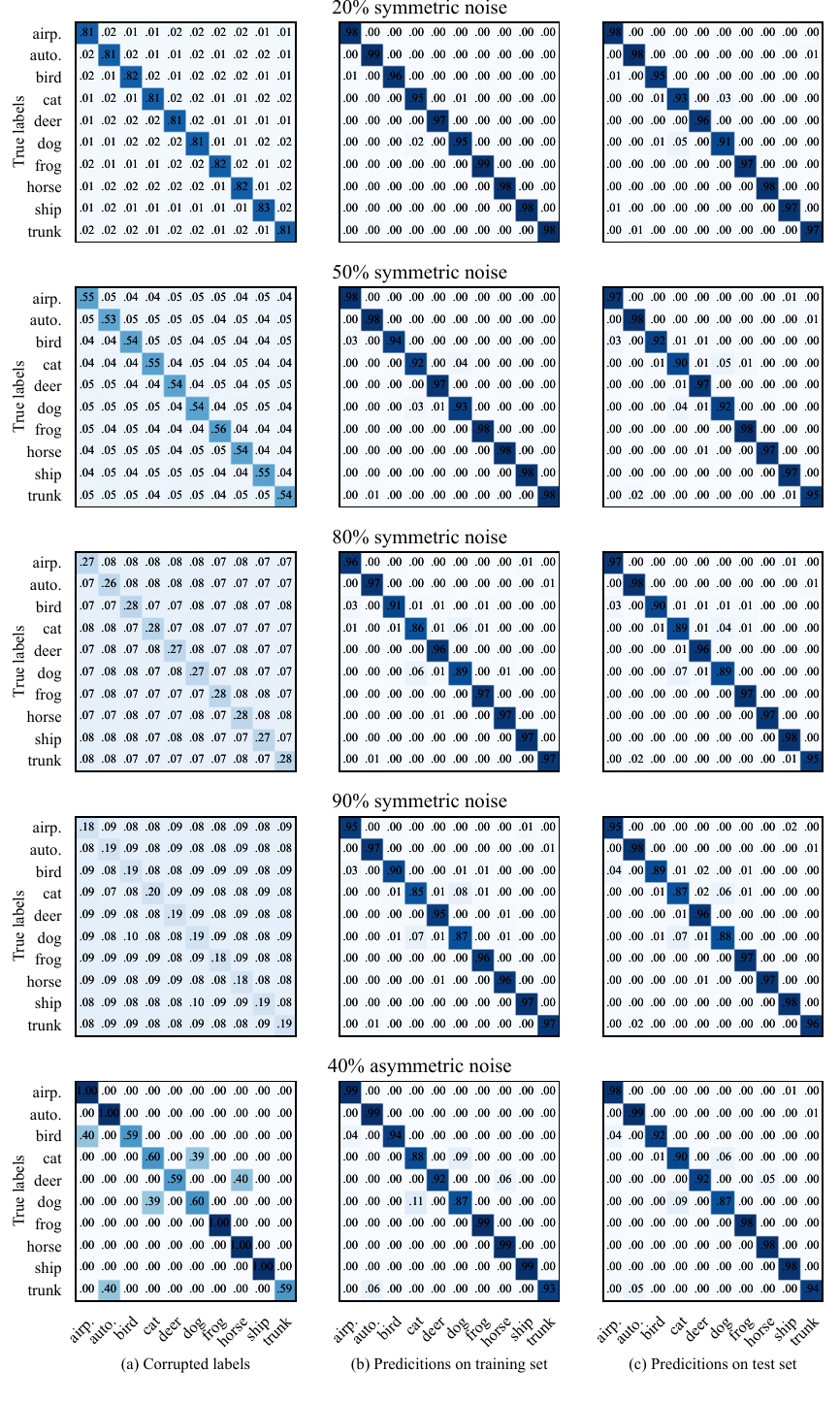} 
    \caption{Confusion matrices on CIFAR-10 under 20\%-90\% symmetric noise and 40\% asymmetric noise. 
    (a) The corrupted training set.
    (b) The prediction on the training set.
    (c) The prediction on the test set.
    The airp. and auto. are airplane and automobile for short.} 
    \label{fig_confison}
\end{figure}

\subsubsection{Benchmark Datasets} 
We benchmark the proposed method on experimental settings using CIFAR-10, CIFAR-100 \cite{krizhevsky2009learning} with different levels of synthetic noises, as well as the real-world dataset Mini-WebVision\cite{li2017webvision}.
On CIFAR-10 and CIFAR-100, There are two commonly used types of synthetic noise  \cite{kim2019nlnl,li2020dividemix}: symmetric noise and asymmetric noise.
Symmetric noise corrupts samples to random classes with the same probability, while asymmetric noise corrupts samples to specific classes according to a pre-defined label transition matrix (shown in Fig. \ref{fig_confison}(a)).
The noise rate ranges from 20\% to 90\% (note that samples are randomly corrupted to $C$ classes for symmetric noise, and the true labels may be maintained afterward).
For Mini-WebVision, we use the first 50 classes of the Google image subset. The ImageNet ILSVRC12 is used as the validation set following \cite{chen2019understanding,li2020dividemix}.
    
\subsubsection{Backbone Models} 
The backbone for CIFAR is the 18-layer PreAct Resnet \cite{he2016identity}. 
The backbone for Mini-WebVision is the Inception-ResNet v2 \cite{szegedy2017inception}. 

\subsubsection{Training Schema} 
For all experiments, we mainly tune two hyper-parameters, namely the temperature $T_{pl}$ for pseudo-labeling and the $k$ in the $k$-NN graph. 
Specifically, using a validation set of 5000 samples, we choose $T_{pl}$ from $\{1,2,3,4\}$ and $k$ from $\{2,10,50,100\}$.
For two light noise settings, namely CIFAR-10 under 20\% symmetric noise and CIFAR-100 under 20\% symmetric noise, and Mini-WebVision, the $T_{pl}$ is set to $1$.
Otherwise, the $T_{pl}$ is set to 2.
For CIFAR-10 under symmetric noise, $k$ is set to 50. 
In other cases, the $k$ is set to 2.
For CIFAR-10 and CIFAR-100, the network is trained using SGD with a learning rate of 0.01, a momentum of 0.9, a weight decay of 0.0005, and a batch size of 128 for 400 rounds. 
The model is warmed up for 15 epochs (simple supervised training using the original noisy dataset). 
We reduce the learning rate to 0.001 in the last 100 training rounds.
For Mini-WebVision, the network is trained using SGD with a learning rate of 0.01, a momentum of 0.9, a weight decay of 0.0005, and a batch size of 160 for 300 rounds. 
The warm-up period is 1 epoch. 
We reduce the learning rate to 0.001 in the last 100 training rounds.

We add a regularization term for encouraging the network output uniform distribution following many LNL methods \cite{tanaka2018joint,arazo2019unsupervised,li2020dividemix}:
$L_{reg}=\sum_c \pi_c log(\frac{\pi_c}{\Bar{\text{p}}_c})$ where $\Bar{\text{p}}_c=\frac{1}{B}\sum_{i=1}^B \text{p}(y=c\mid x_i;\theta)$. $\pi$ is the uniform prior distribution, we set $\pi_c=\frac{1}{C}$. 
For asymmetric noise, we add a negative entropy loss term during warm-up following \cite{pereyra2017regularizing,li2020dividemix}: $L_{asym}=\sum_c \text{p}(y\mid x;\theta)log(\text{p}(y\mid x;\theta))$.

\begin{table}
    \centering
    \caption{
        List of operations for strong transformations of the modified RandAugment. 
        Three transformations are randomly chosen and performed with stochastic magnitude.
    }
    \label{strongaug}
    \begin{tabular}{ll|ll}
        \toprule
        Operation      & Range       &Operation      & Range         \\ \midrule
        % CIFAR-10,CIFAR-100 
        AutoContrast   & [0, 1]      &Rotate         & [-30, 30]    \\
        Brightness     & [0.05, 0.95]&Sharpness      & [0.05, 0.95] \\
        Color          & [0.05, 0.95]&ShearX         & [-0.3, 0.3]  \\
        Contrast       & [0.05, 0.95]&ShearY         & [-0.3, 0.3]  \\
        Equalize       & [0, 1]      &Solarize       & [0, 256]     \\
        Identity       & [0, 1]      &TranslateX     & [-0.3, 0.3]  \\
        Posterize      & [4, 8]      &TranslateY     & [-0.3, 0.3]  \\
        \bottomrule
        % Cut            & []           \\  
    \end{tabular}
\end{table}

\subsubsection{Data augmentation} 
\label{Details of transformations}

For the data augmentation $\mathcal{A}$, we use the modified version of RandAugment \cite{cubuk2020randaugment} follows the setting of FixMatch \cite{sohn2020fixmatch}.
The operations of RandAugment are shown in Table~\ref{strongaug}.
The meaning of range is the same as the original version, so we don't elaborate here.
% \begin{table}
%     \centering
%     \caption{Implementation details}
%     \label{table_detail}
%     \begin{tabular}{l|cc|cc}
%         \toprule
%     \end{tabular}
% \end{table}

\subsection{Comparison to SOTA}
\subsubsection{CIFAR-10, CIFAR-100}
\label{sec:cifar}

For comparison on CIFAR-10 and CIFAR-100, results of Bootstrap \cite{reed2014training}, F-correction \cite{patrini2017making}, P-correction \cite{yi2019probabilistic}, M-correction \cite{arazo2019unsupervised} Mixup \cite{zhang2017mixup}, Co-teaching+ \cite{yu2019does}, Meta-learning \cite{li2019learning}, DivideMix \cite{li2020dividemix},  AugDesc \cite{nishi2021augmentation} are reported. % , , 
Their results are from \cite{li2020dividemix,nishi2021augmentation}.
Following \cite{li2020dividemix,nishi2021augmentation}, we report both the \textbf{Best} test accuracy across all epochs and the average test accuracy over the \textbf{Last} 10 epochs.
The performance of LaplaceConfidence over 3 trials with different random seeds for noise generation and parameter initialization is averaged.
It is also worth noting that AugDesc is extended from DivideMix by adding different augmentations on top of it. 
We include two versions of AugDesc: one is with RandAugment \cite{cubuk2020randaugment}, which is the same as the augmentation we use, and another is with AutoAugment \cite{cubuk2019autoaugment}, which uses reinforcement learning to determine the selection and ordering of a set of augmentation functions.

The proposed method outperforms the previous best method by up to 2.8\% on CIFAR-10 under symmetric noise and by up to 14\% on CIFAR-100 under symmetric noise, as shown in Table \ref{table_CIFAR}.
Only AugDesc-AutoAugment achieves competitive results on CIFAR-100 under 50\% noise, with AutoAugment that has a higher computation cost as shown in \cite{cubuk2020randaugment}.
The performance gain is bigger under heavy noise.
We remark that suboptimal confidence estimation would misguide the training, and the model, in turn, overfits the wrong labels and adversely affects the subsequent confidence estimation.
Thus, a good label confidence estimation method could bring huge improvements under heavy noise.

In terms of asymmetric noise, the proposed method also achieves better results, surpassing the previous best method by over 1.8\%.
After training, the model is less biased towards the given class-dependent noise on the training and test set, as shown in Fig. \ref{fig_confison}(b) and (c).

\begin{table}
    \centering
    \caption{Comparison with state-of-the-art methods on Mini-WebVision with real-world noise. The best results are indicated in bold.}
    \label{table_webvision}
    \begin{tabular}{l|cc|cc}
        \toprule
        \multirow{2}{*}{Method} & \multicolumn{2}{c|}{Mini-WebVision} & \multicolumn{2}{c}{ILSVRC12} \\ \cmidrule(r){2-3}\cmidrule(l){4-5}
                                & top-1  & top-5  & top-1  & top-5  \\ \midrule
        % F-correction            & 61.12 & 82.68 & 57.36 & 82.36 \\
        % Decoupling              & 62.54 & 84.74 & 58.26 & 82.26 \\
        D2L                     & 62.68 & 84.00 & 57.80 & 81.36 \\
        MentorNet               & 63.00 & 81.40 & 57.80 & 79.92 \\
        Co-teaching             & 63.58 & 85.20 & 61.48 & 84.70 \\
        Iterative-CV            & 65.24 & 85.34 & 61.60 & 84.98 \\
        DivideMix               & 77.32 & 91.64 & 75.20 & 90.84 \\ 
        NGC                     & 79.16 & 91.84 & 74.44 & 91.04 \\ 
        LaplaceConfidence       & \textbf{80.52} & \textbf{94.56} & \textbf{77.36} & \textbf{94.12} \\ \bottomrule
    \end{tabular}
\end{table}

\begin{figure*}[h]
    \centering
    \includegraphics[width=0.95\textwidth]{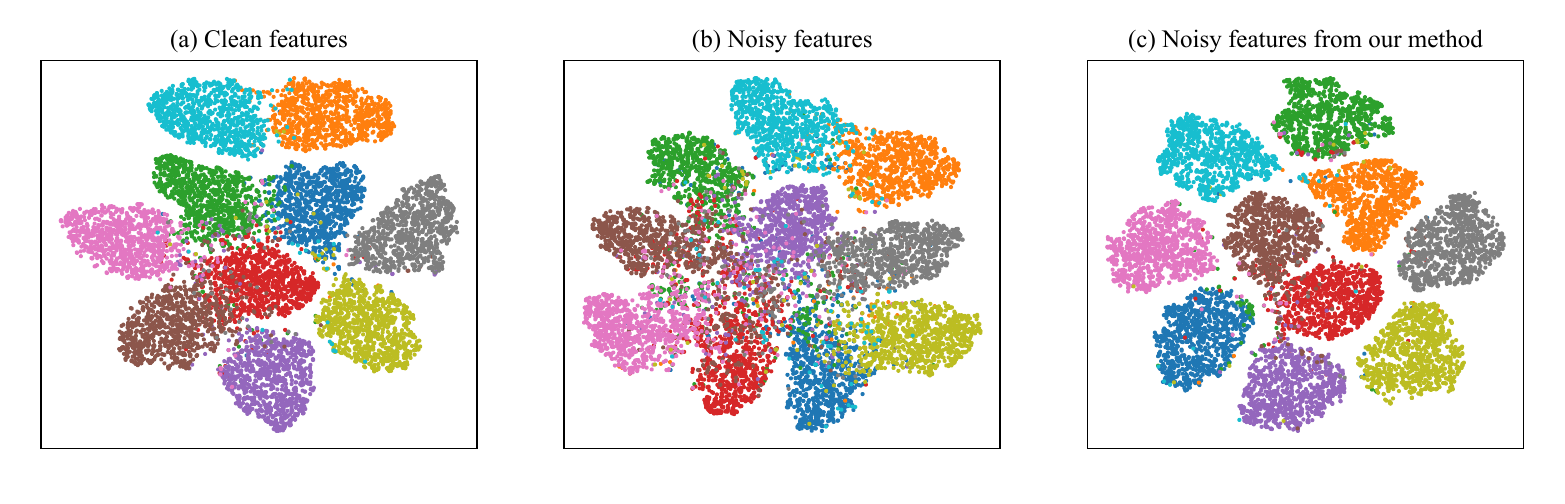} 
    \caption{%DivideMix's (a) test accuracy and (b) quality of confidence estimation during training.
    % (a) and (b) show the overlap of loss distribution on CIFAR-10 under different noise levels. 
    Feature representations visualized by t-SNE.
    Color represents the ground-truth class.
    (a). The clean feature representation (trained on clean CIFAR-10).  
    (b). The noisy feature representation (trained on 50\% uniformly corrupted CIFAR-10 without robust training techniques).  
    (c). The noisy feature representation our method learned.
    The standard softmax-based linear classifier could easily fail on the twisted feature in (b), yielding bad label confidence estimation.  
    Best viewed in color.
        }
    \label{feature_distribution}
\end{figure*}

In Fig. \ref{feature_distribution}(a) and (b), we visualize the learned features of clean and noisy datasets.
It can be seen that clean features have a clear cluster structure, where samples in the same class are close.
However, noisy features are more twisted, as in Fig. \ref{feature_distribution}(b), where some samples locate far away from their cluster center. 
The softmax-based linear classifier on top of the penultimate layer features would be unable to separate such twisted features and, subsequently, the cross-entropy loss is unable to identify label noise \cite{DBLP:conf/cvpr/IscenVAS22,DBLP:conf/nips/KimKCCY21}.
We suggest that this is one of the reasons why the small-loss criterion fails.

From Fig. \ref{feature_distribution} (b), we can also notice that most samples, even though they do not form cluster structure, still have same-class neighbors.  
Therefore, we attempt to utilize the topological relationships between samples for label confidence estimation.
The possibility of a sample/node being mislabeled can be determined by all data points' distribution in the feature space.
The graph structure is commonly used to model such relationships, on which all nodes can propagate their labels to their neighbors.
Considering that the propagation could be an iterative process, where corrected nodes can affect their neighbors again until reaching convergence, we solve the label confidence estimation as the classic graph Laplacian minimization problem \cite{newman2004detecting,DBLP:conf/icml/ZhuGL03,DBLP:journals/corr/abs-2106-04527}, i.e., making the intrinsic connection structure sufficiently smooth.
After this process, the more the label changes, the more possible it is that the original labels are wrong.

\subsubsection{Mini-WebVision}

Learning with real-world label noise is more challenging.
We evaluate  LaplaceConfidence on Mini-WebVision to verify that it performs well on a larger dataset with more complex noise. 
For comparison, we choose classification-loss-based methods, namely MentorNet \cite{jiang2018mentornet}, Co-teaching \cite{han2018co}, Iterative-CV \cite{chen2019understanding}, DivideMix \cite{li2020dividemix}, and feature-based methods, namely D2L \cite{ma2018dimensionality} and NGC \cite{wu2021ngc}. 
It is worth noting that NGC is also and graph-based LNL method which utilizes the largest connected components for noise identification. 
Comparison with it can verify our method based on Laplacian energy minimization is better than other graph-based methods.

LaplaceConfidence outperforms all other classification-loss-based and feature-based LNL methods by achieving a top-1 and top-5 accuracy of 80.52\% and 94.56\%, respectively.
It is 1.35\% and 2.71\% better than the previous best NGC, respectively.

\subsection{Accelerating LaplaceConfidence}
% \subsection{Time Complexity Analysis}
The computation time difference between LaplaceConfidence and conventional loss-based LNL methods is due to the label confidence estimation process.
Given that the field of LNL often deal with large datasets or models, reducing the computation cost further would be beneficial for the potential real-world usage of our method.
Since our method takes a bilevel optimization form, and we do not need to back-propagate the gradient in the label confidence process, the features can be freely manipulated.
Therefore, we consider a simple strategy that is known to be capable of removing irrelevant or redundant features, i.e., reducing the dimension of the extracted feature embeddings using Principal Component Analysis (PCA) for the calculation in Eq. \ref{E normalized adjacency matrix}.

\begin{table}
    \caption{Time cost in seconds and accuracy. ``per conv.'' stands for the running time of per convergence of the estimation method.}%, and ``per epoch'' stands for the training time per epoch, which includes the confidence estimation process and the model training process.
    \label{table_time}
    \centering
    % \resizebox{.5\textwidth}{!}{%
    \begin{tabular}{cccc}%{ll|S[table-format=3.2]S[table-format=3.2]S[table-format=3.2]}
        \toprule
         & &  LC & {PCA+LC} \\ \midrule
     \multirow{2}{*}{CIFAR-10}      & Time cost per conv.       & 8.66    &  \textbf{5.66} \\                 %& 0.07    
                                    % & Time cost per epoch       & 70.16   & 66.82  \\                %& 57.55   
                                    & Average accuracy          & \textbf{95.45}    &  95.41  \\ \midrule     %& 94.91   
     \multirow{2}{*}{Mini-WebVision}& Time cost per conv.       & 22.89 &     \textbf{14.53}   \\             %& 0.07    
                                    % & Time cost per epoch       & 1495.39  & 1481.41  \\             %& 1438.41 
                                    & Average accuracy           & 80.52  &    \textbf{81.32}    \\ \midrule  %& 57.55   
    \end{tabular}
    % }
    % }
\end{table}

To further accelerate the process of $k$-NN graph construction, we implement it using the Faiss library (\url{https://faiss.ai/}).
Even though the underlying $k$-selection algorithm's worst time complexity is still $\mathcal{O}(n^2)$, the average case is reduced to $\mathcal{O}(n)$.
What's more, the Faiss library provides quick GPU implementation and optimization, which takes advantage of parallelism.
On the same infrastructure, we test the running time of LaplaceConfidence and LaplaceConfidence+PCA on two datasets.

% To solve the linear system, we use the conjugate gradient approach.
% We follow the standard implementation, and its time complexity is $\mathcal{O}(n)$ per iteration.

% \begin{table*}[htbp]
%     \centering
%     \begin{tabular}{lr|c|c|c|c|c||c|c|c|c}
%         \toprule
% 		Dataset               &      &\multicolumn{5}{c||}{CIFAR-10}& \multicolumn{4}{c}{CIFAR-100}\\\midrule%\cmidrule(r){1-2}\cmidrule(Robust LR){3-7}\cmidrule(l){8-11}
% 		Noise type &      &\multicolumn{4}{c|}{Sym.}& \multicolumn{1}{c||}{Asym.} & \multicolumn{4}{c}{Sym.}\\\midrule
% 		\multicolumn{2}{l|}{Method/Noise ratio}        & 20\% & 50\% & 80\% & 90\% & 40\% & 20\% & 50\% & 80\% &  90\% \\ \midrule%\cmidrule(r){1-2}\cmidrule(Robust LR){3-7}\cmidrule(l){8-11}
% 		\multirow{2}{*}{LaplaceConfidence}             & Best & \textbf{96.4} & \textbf{96.0} & \textbf{95.0} & \textbf{94.7} & \textbf{95.2} & \textbf{79.6} & 76.5 & \textbf{70.4} &  \textbf{55.2} \\
% 				              & Last & \textbf{96.3} & \textbf{95.8} & \textbf{94.8} & \textbf{94.6} & \textbf{94.7} & \textbf{79.3} & 75.5 & \textbf{69.4} & \textbf{44.6} \\
% 		\multirow{2}{*}{LaplaceConfidence-PCA}          & Best & \textbf{96.2} & \textbf{96.0} & \textbf{95.3} & \textbf{94.6} & \textbf{94.6} & \textbf{79.4} & 76.8 & \textbf{68.2} &  \textbf{38.8} \\
% 				                                    & Last & \textbf{96.2} & \textbf{95.8} & \textbf{95.1} & \textbf{94.1} & \textbf{94.4} & \textbf{78.9} & 76.0 & \textbf{67.2} & \textbf{37.1} \\
%         \bottomrule
%     \end{tabular}
%     \caption{PCA}
%     \label{table_PCA}
% \end{table*}

We run our experiments and calculate the computational cost on 24 cores Intel(R) Xeon(R) Platinum 8255C CPU and a single NVIDIA RTX V100 GPU to get the real time cost.
As shown in Table \ref{table_time}, by simply adding the PCA technique, LaplaceConfidence is quicker with almost no accuracy loss on CIFAR-10.
Noteworthy, the PCA even improves the accuracy on Mini-WebVision by 0.8\%.
We remark that it is because the dimension reduction, keeping the important features in the learned representation, could reduce the damage of ambiguous/wrong features learned from the noisy training signals.

\begin{table*}[t]
    \caption{Ablation study. Results on CIFAR-10 with different levels of symmetric noise are reported.}
    \label{table_ab}
    \centering
    \begin{tabular}{ll|cccc}
        \toprule
		\multicolumn{2}{l|}{Method/Noise ratio}                           & 20\% & 50\% & 80\% & 90\%   \\ \midrule
		
		\multirowcell{2}[0ex][l]{LaplaceConfidence}            & Best & \textbf{96.4} & \textbf{96.0} & \textbf{95.0} & \textbf{94.7}  \\
				                                               & Last & \textbf{96.3} & \textbf{95.8} & \textbf{94.8} & \textbf{94.6}  \\ \midrule
		\multirowcell{2}[0ex][l]{Replacing LaplaceConfidence with GMM}    & Best & 96.4 & 95.6 & 94.8 & 93.4   \\
		                                                                  & Last & 96.2 & 95.5 & 94.3 & 93.1   \\ \midrule
% 		\multirowcell{2}[0ex][l]{Without RandAugment}                     & Best & 93.8 & 89.5 & 76.7 & 57.9 \\ %
% 		                                                                  & Last & 92.8 & 77.5 & 76.4 & 57.4 \\ \midrule
		\multirowcell{2}[0ex][l]{Without co-traininng}                    & Best & 96.0 & 95.3 & 93.9 & 94.0 \\ %
		                                                                  & Last & 95.4 & 95.2 & 93.3 & 93.6 \\  
        \bottomrule
    \end{tabular}
\end{table*}

\begin{figure*}
    \centering
    \includegraphics[width=\textwidth]{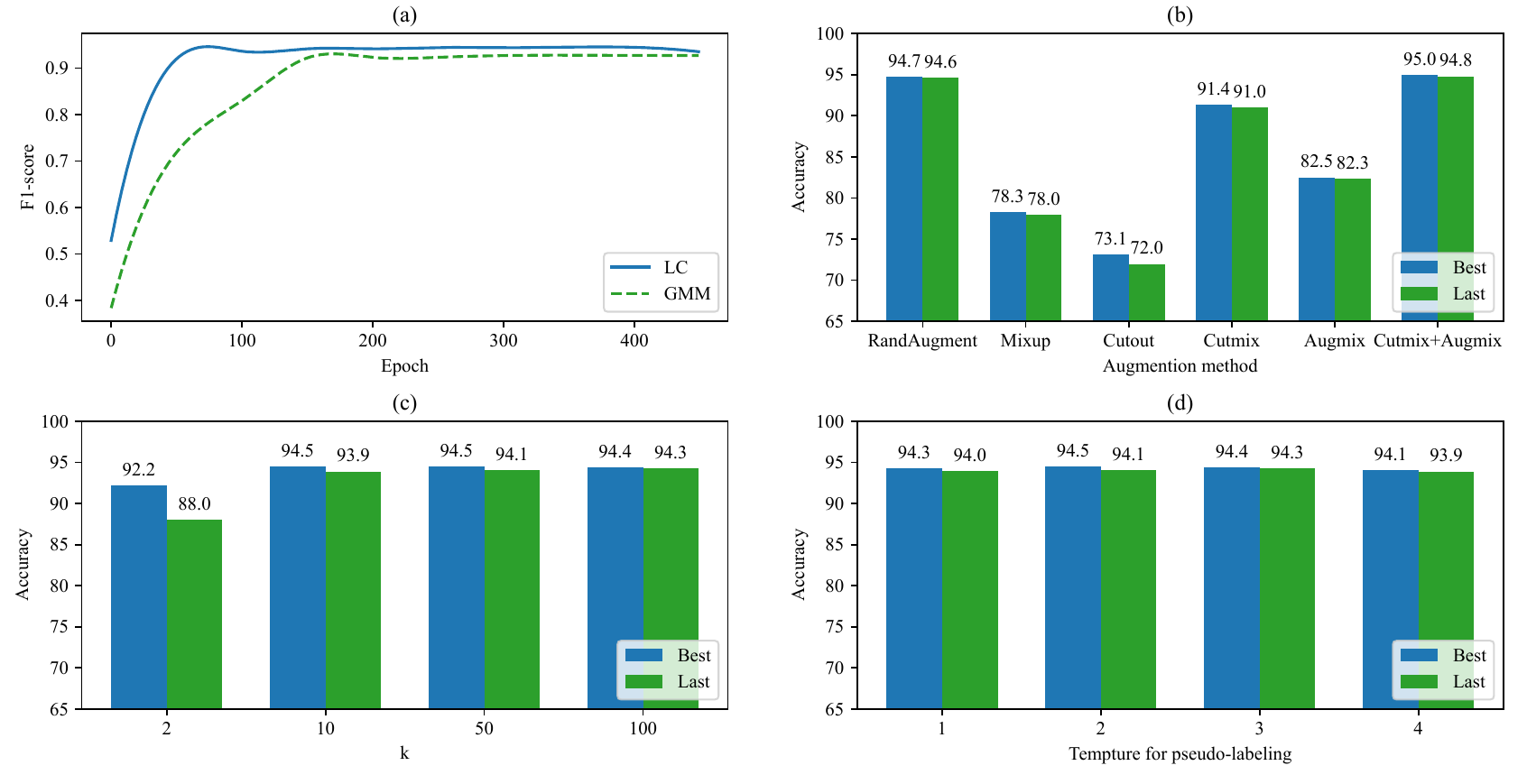} 
    \caption{(a) The quality of confidence estimation of LaplaceConfidence and GMM on CIFAR-10 under 90\% noise. (b) LaplaceConfidence with other augmentations on CIFAR-10 under 90\% noise. (c) Varying the $k$ for $k$-NN graph. (d) Varying the temperature $T_{pl}$ for pseudo-labeling. In (a), a threshold of 0.5 is used for separating clean and noisy. F1-score is used considering the imbalance between clean and noise. LC is short for LaplaceConfidence.} 
    \label{fig_gmmcvlc}
\end{figure*}

\subsection{Ablation Study}

To study the importance of the main components in LaplaceConfidence, we test each of them separately and report the performance:
% \begin{itemize}
\begin{itemize}
    \item To study the effect of LaplaceConfidence, we replace it with GMM. 
    \item To study the effect of the co-training, we only use one model.
    \item To study the effect of the augmentation, we replace RandAugment with other augmentation methods.
\end{itemize} %[1.]
We also study the influence of two important hyper-parameters in Fig. \ref{table_ab}, namely the number of nearest neighbors $k$ and the temperature $T$.
% We also show how the $k$ ()  affects the performance in Fig. \ref{fig_gmmcvlc}(c). 
% The gain becomes insignificant after 50. 

\subsubsection{LaplaceConfidence} 
LaplaceConfidence plays a key role in our method. 
For comparison, we choose Gaussian Mixture Model (GMM), the confidence estimation method used in the previous SOTA.
A mixture distribution of two Gaussian distributions is fitted on the loss value using the Expectation-Maximization algorithm.
Regarding its hyper-parameters setting and implementation, we follow the official implementation of DivideMix (\url{https://github.com/LiJunnan1992/DivideMix}).
As shown in Table \ref{table_ab}, the accuracy of LaplaceConfidence is better than GMM.
We further compare their confidence estimation quality during training in Fig. \ref{fig_gmmcvlc}(a), which confirms LaplaceConfidence's superiority against GMM (we use F1 score because the clean-noisy binary classification is imbalanced).
What is more, we find that the estimation of GMM is unstable under heavy noise.
It is because the loss of clean and noisy samples seriously overlaps, and the GMM fails to converge under heavy noise.
Note that DivideMix models the averaged loss over the last 5 epochs to improve convergence stability.
We also add it for the experiment of GMM. Otherwise, the model would collapse.
The proposed LaplaceConfidence, on the other hand, produces better estimation and does not need extra tricks to stabilize training.

% We also show how the $k$ (the number of nearest neighbors)  affects the performance in Fig. \ref{fig_gmmcvlc}(c). 
% The gain becomes insignificant after 50. 
%*********

% Considering the imbalance between the number of clean samples and noisy samples under heavy noise, we also include F1-score. 
% Note that we use the outputs of our model to calculate both LaplaceConfidence and GMM. 
% Therefore, the gaps between accuracy and F1-score show \textbf{LaplaceConfidence outperforms GMM by a small margin in every estimation based on the same model}.
% When only the confidence estimation from GMM is used to subsequent training, the final model is significantly worse, as in Table \ref{table_ab}.

\subsubsection{Co-training} 
Co-training brings performance gain because it alleviates the error accumulation problem in the self-training process and aggregates the predictions of two models.
We find that co-training almost always outperforms one single model. 
However, it introduces extra computational costs (more than doubled) and design choices.

\subsubsection{Data augmentations}
Augmentation has been found useful in many tasks, such as semi-supervised learning \cite{sohn2020fixmatch}, and unsupervised learning \cite{van2020scan,khosla2020supervised}.
We report LaplaceConfidence with other augmentations methods in Fig. \ref{fig_gmmcvlc} (b).
One may argue that data augmentation has such a big influence that it could cause unfair comparisons.
We show that LaplaceConfidence outperforms other methods that use the same or even stronger augmentation as in Table \ref{table_CIFAR}.
It also does not diminish the contributions of other components because they further make improvements upon data augmentation, as in Table \ref{table_ab}.
We find that LaplaceConfidence with Cutmix+Augmix achieves the best and the last accuracy of 95.0\% and 94.8\% on CIFAR-10 under 90\% noise, respectively. 
Considering that no previous LNL method use Augmix, we only report the RandAugment version of LaplaceConfidence for a fair comparison.

\section{Conclusion}
This paper studies the key problem in LNL: label confidence estimation. 
We propose LaplaceConfidence, a new graph-based method that utilizes the rich topological information in the feature space.
It is superior to previous feature-based methods by correcting the bias in the label confidence estimation caused by mislabeled data points.
We demonstrate that our approach beats other methods through systematical experiments, significantly advancing the state-of-the-art.
We also find that reducing the dimension of learned features before calculating the feature similarities permits smaller computations without damaging generalization.
Furthermore, we conduct ablation experiments to study the effects of our components.

There are many possible avenues for future research into label confidence estimation, including the exploration of other forms of noise present in datasets, such as distribution shifted data \cite{DBLP:conf/cvpr/LeeHZY18} or out-of-distribution data \cite{DBLP:conf/cvpr/WangLMBZSX18}. With LaplaceConfidence as a starting point, we believe that techniques from graph-based approaches may be adapted to solve these more challenging problems. For instance, one could adjust the contribution of different examples by introducing different weights for nodes in the graph.
Due to its flexibility and scalability, we anticipate that our method can be applied to a range of real-world applications.